\documentclass[10pt,twocolumn,letterpaper]{article}

\usepackage{cvpr}              
\usepackage[accsupp]{axessibility}

\definecolor{cvprblue}{rgb}{0.21,0.49,0.74}
\usepackage[pagebackref,breaklinks,colorlinks,allcolors=cvprblue]{hyperref}

\usepackage{siunitx}

\def\paperID{****} 
\def\confName{CVPR}
\def\confYear{2026}

\title{3D Gaussian Splatting for Annular Dark Field Scanning Transmission Electron Microscopy Tomography Reconstruction}

\author{
    Beiyuan Zhang\textsuperscript \quad
    Hesong Li\textsuperscript \quad
    Ruiwen Shao\textsuperscript \quad
    Ying Fu\textsuperscript{$\dagger$} \\ 
    \ Beijing Institute of Technology \\ 
    {\tt\small \{byzh,lihesong2,rwshao,fuying\}@bit.edu.cn} 
}

\usepackage{caption}
\usepackage{multirow}

\begin{document}
\maketitle

{
\let\thefootnote\relax\footnotetext{$^\dagger$ Corresponding author.}
}

\begin{abstract}
Analytical Dark Field Scanning Transmission Electron Microscopy (ADF-STEM) tomography reconstructs nanoscale materials in 3D by integrating multi-view tilt-series images, enabling precise analysis of their structural and compositional features. Although integrating more tilt views improves 3D reconstruction, it requires extended electron exposure that risks damaging dose-sensitive materials and introduces drift and misalignment, making it difficult to balance reconstruction fidelity with sample preservation. In practice, sparse-view acquisition is frequently required, yet conventional ADF-STEM methods degrade under limited views, exhibiting artifacts and reduced structural fidelity. To resolve these issues, in this paper, we adapt 3D GS to this domain with three key components. We first model the local scattering strength as a learnable scalar field, denza, to address the mismatch between 3DGS and ADF-STEM imaging physics. 
Then we introduce a coefficient $\gamma$ to stabilize scattering across tilt angles, ensuring consistent denza via scattering view normalization. Finally, We incorporate a loss function that includes a 2D Fourier amplitude term to suppress missing wedge artifacts in sparse-view reconstruction. Experiments on 45-view and 15-view tilt series show that DenZa-Gaussian produces high-fidelity reconstructions and 2D projections that align more closely with original tilts, demonstrating superior robustness under sparse-view conditions.

\end{abstract}    
\section{Introduction}
\label{sec:intro}
Nanoscale materials science relies heavily on advanced characterization techniques to decode the relationship between 3D structure and functionality~\cite{miao2016atomic, lemberg2012mo, oh2025correlative, lee2021single}. Among these, ADF-STEM has become indispensable for studying nanoscale materials, offering atomic-level resolution~\cite{pennycook2011scanning, andersen1984simultaneous}. This unique capability enables visualization of critical nanoscale features, such as core-shell nanoparti-

\begin{flushright}

  \begin{figure}[h!]
    \includegraphics[width=\linewidth]{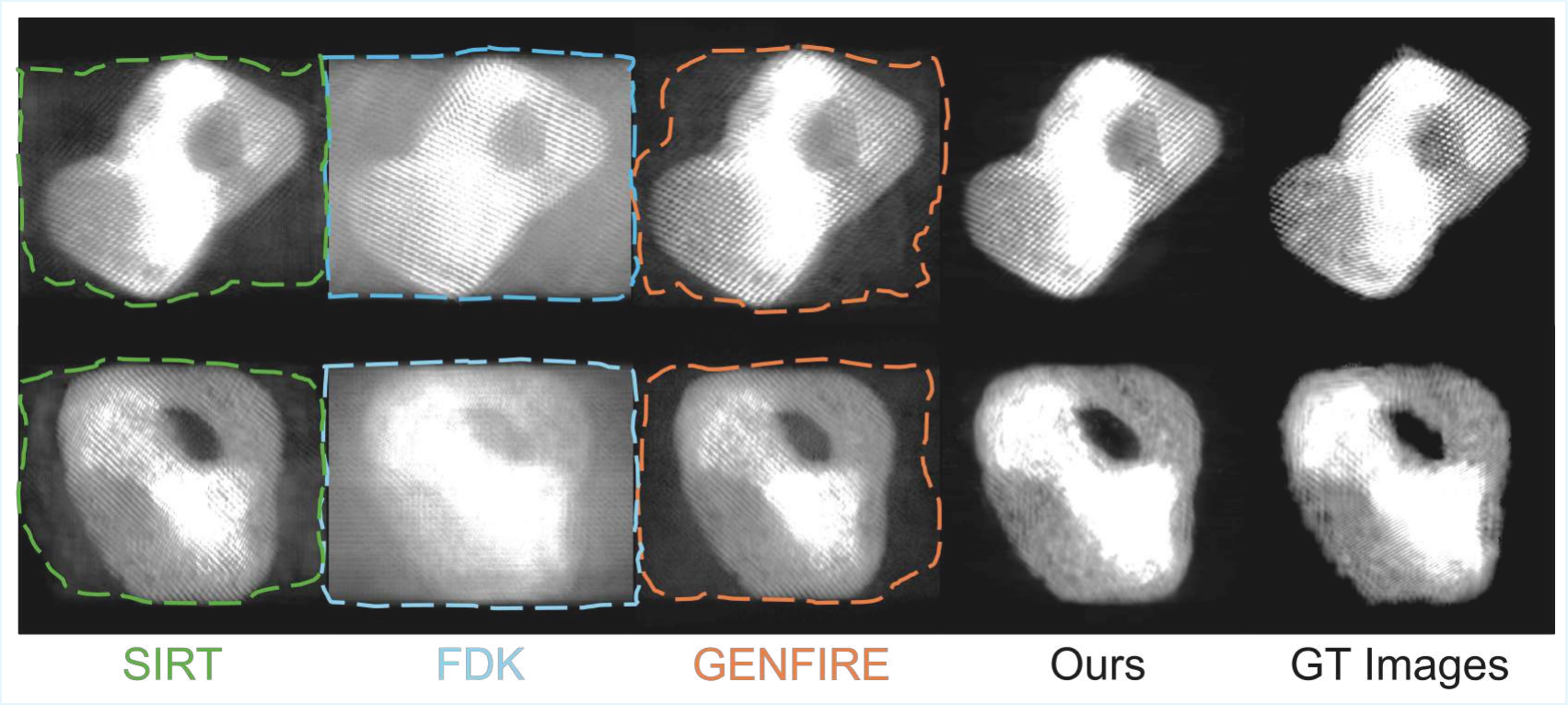}
    \caption{Projection comparison under 15-view reconstruction across traditional reconstruction methods and our approach. Brightness is enhanced while contrast is reduced to highlight artifacts. FDK exhibits the most severe artifacts, followed by SIRT and GENFIRE. Our method yields the cleanest projections.}
    \label{fig:teaser}
  \end{figure}
\end{flushright}

\noindent cle morphologies, pore distributions in catalytic supports, and phase segregation in nanocomposites~\cite{lemberg2012mo, sidky2008image, ube2024three}.

To fully exploit this potential, ADF-STEM tomography reconstructs 3D structural and compositional distributions from 2D projections acquired at varying tilt angles, uncovering spatial heterogeneities that drive material performance~\cite{chen2013three, pryor2017genfire}. 
However, delicate nanomaterials cannot withstand dense tilt-series acquisitions due to their high dose sensitivity. 
Due to the risk of beam-induced damage, dose-sensitive materials cannot tolerate extensive tilt-series acquisition. Consequently, sparse-view imaging is commonly adopted, despite its inherent reconstruction limitations~\cite{leary2013compressed, zha2022naf}.
This severely limits the application of ADF-STEM tomography in characterizing such materials, while tradeoffs between reconstruction speed and fidelity further exacerbate this limitation.
Traditional ADF-STEM tomography workflows struggle to address these challenges simultaneously. 
Analytical methods~\cite{kak2001principles, feldkamp1984practical, radon2007determination} introduce severe missing wedge artifacts when projections are sparse.
Iterative methods~\cite{hageman2012applied, chun2014alternating, sorzano2017survey, pryor2017genfire} reduce artifacts but are computationally cumbersome and often over-smooth fine nanoscale details.

3D Gaussian Splatting (3DGS)~\cite{kerbl20233d} has emerged as a transformative technique in view synthesis, offering real-time rendering via parallelized rasterization and outperforming neural fields in speed and flexibility. To address the limitations of ADF-STEM tomographic reconstruction mentioned above, we adapt 3D Gaussian ADF-STEM tomography and present DenZa-Gaussian for high-quality, efficient and robust reconstruction.

We first reconcile the mismatch between 3DGS and ADF-STEM imaging physics by grounding its design in Rutherford elastic scattering~\cite{pennycook2011scanning}. This rule explicitly encodes Z-contrast as proportional to the product of local density and $Z^\alpha$ that governs electron scattering in ADF-STEM. We simplify $density \cdot Z^\alpha$ into a learnable scalar field, denoted as denza, enabling a closed-form forward model consistent with ADF-STEM imaging geometry.
Then, we introduce a Scattering-View-Consistent coefficient $\gamma$ to ensure cross-view consistency of this product. This modification eliminates inconsistency from standard 3DGS’s neglect of covariance-related scaling during 3D-to-2D projection, preserves the true scattering power of each nanoscale region across all tilt angles. Finally, loss is computed between the projected images and ground truth, along with regularization on the 3D volume. While pixel-wise losses enforce local grayscale consistency, they fail to constrain global frequency-domain structure, witch is crucial for suppressing missing wedge artifacts in sparse-view tomography. To resolve this, we incorporate a 2D Fourier amplitude loss that promotes spatial-frequency coherence, enhancing DenZa-Gaussian’s robustness to sparse projections and preserving fine morphological details.

Experiments on real tilt series with 45-view and 15-view show that DenZa-Gaussian outperforms traditional analytical methods, traditional iterative methods and 3D Gaussian splatting in reconstruction quality. As illustrated by the projection images in Figure.\ref{fig:teaser}, traditional methods introduce noticeable artifacts to varying extents. 3D Gaussian splatting proves inadequate for this task. In comparison, our approach produces clean projections with smooth edges and minimal artifacts, demonstrating the superiority of our method under both standard and sparse-view conditions.

In summary, our main contributions are that we

1.Adapt 3D Gaussian Splatting to ADF-STEM tomography reconstruction, simulate the scattering process of a conical electron beam during projection, derive and simplify the scattering integral formulation, and introduce view-dependent normalization coefficient to enforce consistency of scattering strength across different projection angles.

2.Combine pixel-domain and frequency-domain losses with 3D total variation regularization, suppress artifacts in the reconstructed volume.

3.Conduct experiments and discussion, validate the effectiveness and necessity of our approach, highlight avenues for future research.

\section{Related Work}
\label{sec:related}
In this section, we provide an overview of the related work, including ADF-STEM tomography reconstruction and 3D Gaussian Splatting.

\begin{figure*}[t]
  \centering
  \includegraphics[width=0.85\textwidth]{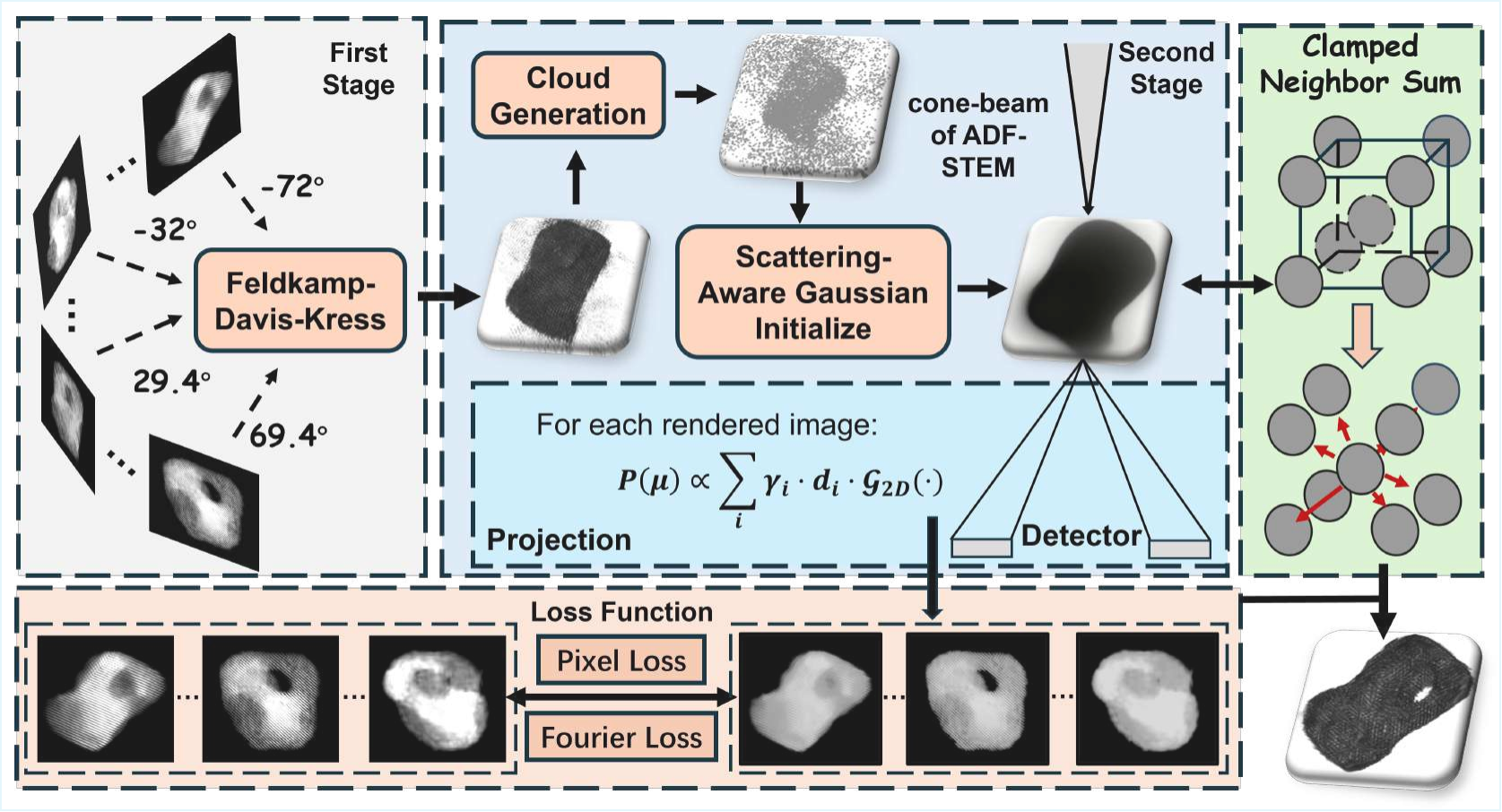}
  \caption{Pipeline of Denza-Gaussian. DenZa-Gaussian is a two-stage Architecture tailored for sparse-view ADF-STEM tomography. The first stage uses FDK initialization to rapidly generate a rough 3D volume capturing the sample’s overall structural topology. The second stage converts this rough volume into a point cloud for Gaussian initialization. We use a custom 2D rendering pipeline optimized for ADF-STEM’s cone-beam imaging geometry to project the initialized 3D Gaussian cloud into 2D images. Loss is computed between these projections and ground truth images, complemented by regularization constraints on the 3D volume. Iterative optimization delivers a high-quality 3D reconstruction.}
  \label{fig:pipeline}
\end{figure*}

\noindent\textbf{ADF-STEM tomography reconstruction.}ADF-STEM tomography reconstruction originated from SIRT/ART iterative schemes developed to suppress missing-wedge artifacts, and was later refined to achieve sub-\AA ngstr\"{o}m lattice imaging under low-dose conditions. 
Subsequent algorithms such as GENFIRE, RESIRE~\cite{pham2023accurate}, CS-based methods~\cite{candes2006robust, donoho2006compressed, wang2020parametric}, and QURT~\cite{baba2020novel} combined oversampled Fourier constraints with real-space positivity, thereby improving noise tolerance and resolution and establishing the approach as a multimodal and multiscale standard. The integration of cryogenic stages, which preserve labile interfaces such as the solid–electrolyte interphase~\cite{wang2018cryogenic}, together with deep learning methods including 3D-Unet~\cite{cciccek20163d} for missing data and U-Net~\cite{ronneberger2015u} for noise filtering, now enables routine three-dimensional mapping of chemistry and valence states in beam-sensitive electrodes such as silicon and catalysts such as PtNi. Coupling with atomic electron tomography allows single-atom three-dimensional coordinate and strain determination with a precision of 15 pm in platinum~\cite{lee2021single}, while in-situ setups capture dynamic processes such as alloying and dislocation motion~\cite{meng2025situ} in four dimensions.

\noindent\textbf{3D Gaussian Splatting.}
3D Gaussian Splatting (3DGS) enables real-time novel-view synthesis by employing explicit anisotropic Gaussians and differentiable rasterization, outperforming NeRF~\cite{mildenhall2021nerf} in both training speed and rendering quality, achieving smooth visual effects even for complex scenes on consumer GPUs.
This representation has been widely adopted across practical scenarios, including dynamic scene modeling that decomposes static and dynamic elements for high-fidelity reconstruction~\cite{wu20244d, zhang2024garfield++}, human pose reconstruction that fuses structural priors for fast and accurate results~\cite{kocabas2024hugs, moreau2024human}, 3D content generation that supports text-driven scene creation~\cite{li2024dreamscene, tang2023dreamgaussian}, and industrial inspection that eliminates pose and illumination sensitivity for defect detection~\cite{liu2024splatpose+}.
Recent works further introduce it into medical imaging, achieving reliable sparse-view CT reconstruction with reduced radiation dose~\cite{zha2024r, yu2025x} and accurate cryo-EM structure recovery~\cite{qu2025gem} that resolves fine-grained molecular conformations, highlighting its significant potential in advancing scientific research.

\section{Method}
In this section, we first introduce our motivation. Then we propose Scattering-Aware Gaussian Splatting, architecture of DenZa-Gaussian and the learning details.
\subsection{Motivation}
ADF-STEM tomography reconstructs 3D nanoscale morphology from multi-view electron projections. However, due to the destructive nature of high-energy electron beams, the number of available views is often severely limited to minimize sample damage, especially for beam-sensitive materials such as biological specimens, polymers, or low-dimensional nanostructures. This constraint not only reduces angular coverage but also exacerbates the missing wedge problem, posing significant challenges for accurate 3D reconstruction. 

Given 3D Gaussian Splatting offers continuous volumetric representation, view-consistent rendering, sub-voxel precision, and efficient and high-fidelity synthesis, we consider it particularly suitable for achieving high-quality reconstruction under sparse-view conditions in ADF-STEM tomography. We bring 3D Gaussian Splatting into this area through physics calibration, Scattering-View-Consistent covariance scaling, combined pixel-wise and Fourier-amplitude losses, alongside 3D total variation regularization.

Physics-informed calibration is rooted in Rutherford elastic scattering, embedding the Z-contrast mechanism into each Gaussian kernel via a unified scalar field, denza. This formulation preserves physical fidelity while remaining tractable under unknown elemental distributions. To account for ADF-STEM’s cone-beam geometry, we introduce a Scattering-View-Consistent coefficient $\gamma$, ensuring each Gaussian’s contribution aligns with the linear integration of scattering along beam paths and remains consistent across tilt angles.

The loss function combines pixel-wise intensity matching with a 2D Fourier amplitude loss, promoting both local detail and global frequency coherence. While pixel losses capture fine grayscale variations, the Fourier term compensates for missing spatial frequencies in sparse-view settings by enforcing structural consistency in the frequency domain.

Finally, a 3D total variation penalty encourages piecewise smoothness in the reconstructed volume and suppresses noise amplification. Together, these components enhance robustness to sparse projections and preserve sub-50 nm morphological details through balanced spatial and frequency-domain constraints.

\subsection{Scattering-Aware Gaussian Splatting}
\label{physic}
To adapt 3D Gaussian Splatting to ADF-STEM tomography, we embed fundamental electron scattering physics into the Gaussian parameterization. Native 3DGS, originally developed for photometric rendering via color and opacity blending, lacks compatibility with the scattering-driven projection mechanisms of ADF-STEM, where image formation arises from electron–sample interactions under a convergent cone-shaped beam. Our scattering-consistent formulation resolves this physical mismatch by aligning Gaussian representation with the principles of Rutherford scattering and cone-beam integration.

This scattering process is driven by a convergent electron beam, which is focused by an electromagnetic lens into a cone-shaped probe defined by its convergence angle and spatial extent. The probe scans point-by-point across the sample surface, and scattered electrons are subsequently collected by an annular dark-field (ADF) detector.
Its scattering intensity follows the Rutherford principle, being proportional to the product of local atomic density and the atomic number Z raised to a power $\alpha$. Here, $\alpha$ is a correction exponent that quantifies the deviation of electron scattering intensity from the ideal $Z^\alpha$ dependence due to electron cloud shielding effects in ADF-STEM imaging. It typically falls in the range [1.3, 2.0], and we set it as a fixed value of 1.7.

ADF-STEM employs a cone-shaped electron beam, and its projection intensity stems from the linear integration of electron scattering intensity along the beam paths. This physical mechanism underpins our framework design. Mathematically, the cone-shaped electron beam in ADF-STEM can be characterized by its convergence angle $\theta$ and probe size $\sigma$, which refers to the full width at half maximum. For a scanning point $\{x,y\}$, on the sample surface, the electron beam propagates along a conical path parameterized by $r(t) = (x, y, z_0) + tn$, where $n$ is the unit direction vector within the cone, $t$ is the path length, and $z_0$ is the initial depth. The scattering intensity at position $r(t)$ is given by 
\begin{equation}
I(r(t)) = k \cdot \rho(r(t)) \cdot Z^\alpha(r(t)),
\end{equation}
where $k$ is a constant related to electron beam energy and experimental setup, $\rho$ is local atomic density, and $\alpha \in [1.3, 2.0]$ as defined. The ADF-STEM projection intensity at $\{x,y\}$ is the integral of $I(r(t))$ along the conical path, \emph{i.e.},
\begin{equation}
P(x,y) = \int_{t_{\text{min}}}^{t_{\text{max}}} I(r(t)) \, dt = k \int_{t_{\text{min}}}^{t_{\text{max}}} \rho(r(t)) \cdot Z^\alpha(r(t)) \, dt.
\label{eq:projection_1}
\end{equation}

Given the unknown elemental distribution of nanomaterials in experiments and the current focus on external morphology reconstruction rather than quantitative elemental analysis, this framework collapses the physical quantity local atomic density times $Z^\alpha$ into a single learnable scalar field termed the denza coefficient, denoted as $d_i$. $d_i$ physically acts as a ``macroscopic proxy" for the integrated scattering capability of the nanoscale volume represented by its corresponding 3D Gaussian. It encapsulates not only the collective effect of atomic number and density but also the spatial extent of the region through the support $\Omega_i$. This makes $d_i$ a direct link between the microscopic atomic arrangement, unobservable in experiments, and the macroscopic ADF-STEM intensity, a measurable signal. A larger $d_i$ indicates a stronger scattering contribution from that region to the final image. With the denza coefficient $d = density \cdot Z^\alpha$, Equation~\eqref{eq:projection_1} simplifies to

\begin{equation}
P(x,y) = k \int_{t_{\text{min}}}^{t_{\text{max}}} d(r(t)) \, dt,
\end{equation}
directly linking the cone-beam geometry to scattering intensity integration.

Each 3D Gaussian is paired with the denza coefficient $d_i$, which directly encodes the effective scattering intensity of the nanoscale region it represents, thereby transforming Gaussians from color carriers in native 3DGS to scattering intensity carriers in ADF-STEM imaging. This design enables the Gaussian point cloud to adaptively match scattering intensity variations across heterogeneous regions, such as the shell and core of core-shell nanoparticles, or pores and matrices in composite materials.

To conform to ADF-STEM’s physical mechanisms and cone-beam geometric constraints, we introduce the Scattering-View-Consistent coefficient $\gamma_i$, a covariance determinant ratio-based scaling factor overlooked by generic 3DGS. It rectifies 3D-to-2D projection scaling discrepancies, ensuring view-invariant scattering contributions from $d_i$ across tilt angles. Unlike native 3DGS, which violates Rutherford scattering intensity conservation, our coefficient explicitly enforces this physical constraint.
Mathematically, this link is manifested in the projection integral as  $\gamma_i$ weighting the product of $d_i$ and the 2D Gaussian projection of the 3D Gaussian, such that,
\begin{equation}
P(\boldsymbol{u}) \propto \sum_i \gamma_i \cdot d_i \cdot \mathcal{G}_{2D}(\cdot),
\end{equation}
which directly translates the physical scattering strength into image contrast.

\subsection{Architecture of Denza-Gaussian}

DenZa-Gaussian is a two stage architecture developed for sparse view ADF STEM tomography. The pipeline is shown in Figure.\ref{fig:pipeline}, and the workflow balances computational efficiency and reconstruction quality by tightly integrating a physics based model of ADF STEM image formation with data driven optimization.

In the first stage, Feldkamp-Davis-Kress(FDK) reconstruction is used to quickly produce a coarse three dimensional volume. This volume captures the specimen overall structural topology and provides a stable starting point grounded in classical tomographic reconstruction rather than a random initialization. Since FDK is computationally efficient, the initialization imposes minimal computational overhead.

The second stage converts the coarse volume into a point cloud that initializes the three dimensional Gaussians. Each Gaussian carries two learned coefficients that make it a physically meaningful scattering intensity element. The first coefficient $d_i$ encodes the integrated scattering capability of the nanoscale region and serves as a learnable proxy that collapses atomic density and effective $Z^\alpha$ as defined in Sec.\ref{physic}. The second coefficient $\gamma_i$ is a normalization factor based on the covariance determinant ratio described in Sec.\ref{physic} and corrects three dimensional to two dimensional projection biases to ensure view invariant scattering contributions across tilt angles. This dual coefficient formulation converts Gaussians from generic color carriers into representations of ADF-STEM scattering intensity. 

We then simulate the cone-beam integration of ADF-STEM by accumulating the weighted contribution of every 3D Gaussian along the electron trajectories, where the weight is the product of $denza$ and $\gamma$. The resulting 2D projections are compared with experimental images through a composite loss that combines the pixel-wise intensity error, the frequency-domain magnitude mismatch, the structural similarity index, and a 3D total-variation penalty to enforce geometric plausibility.

\subsection{Learning Details}

The optimization objective of Denza-Gaussian is formulated as a synergistic combination of data loss terms and regularization terms, separately Corresponding to the lower and right regions in Figure.\ref{fig:pipeline}. These terms are designed to achieve multi-dimensional alignment between rendered projections and experimental measurements while enforcing the structural plausibility of the 3D volume. This integrated design balances data fidelity, which ensures close agreement with observed signals, and structural robustness, which prevents artifacts or overfitting to noise. It does so by leveraging constraints across pixel, frequency, perceptual, and 3D structural domains.

Data loss terms anchor the fitting process across complementary signal domains. The pixel-wise $L_1$ loss enforces baseline intensity consistency via
\begin{equation}
\mathcal{L}_{\text{pixel}} = \frac{1}{N \, |\Omega|}
\sum_{\theta=1}^{N} \left\| I_{\text{render}}^{\theta} - I_{\text{meas}}^{\theta} \right\|_1,
\end{equation} 
where $N$ is the number of projections and $\Omega$ denotes pixel sets. To recover the degraded frequencies of the missing wedge, the 2D Fourier magnitude loss aligns FFT magnitudes using
\begin{align}
\mathcal{L}_{\text{freq}} = \frac{1}{N} 
\sum_{\theta=1}^{N} 
\Biggl[
\frac{1}{|\Psi|} \sum_{(k_u, k_v) \in \Psi} 
\left( 1 + \lambda_{\text{HF}} \cdot w(k_u, k_v) \right) \cdot \notag\\
\quad \left| \left| \mathcal{F}_{\text{render}}^\theta(k_u, k_v) \right| 
- \left| \mathcal{F}_{\text{meas}}^\theta(k_u, k_v) \right| \right|
\Biggr]
\end{align}

\noindent to capture spatial frequency distributions critical to ADF-STEM. The weighting term \( w(k_u, k_v) \) emphasizes high-frequency regions, which are more susceptible to angular undersampling and noise. This encourages recovery of fine structural details that are otherwise suppressed by the missing wedge effect. 

\begin{figure}[t]
  \raggedright
  \includegraphics[width=\columnwidth]{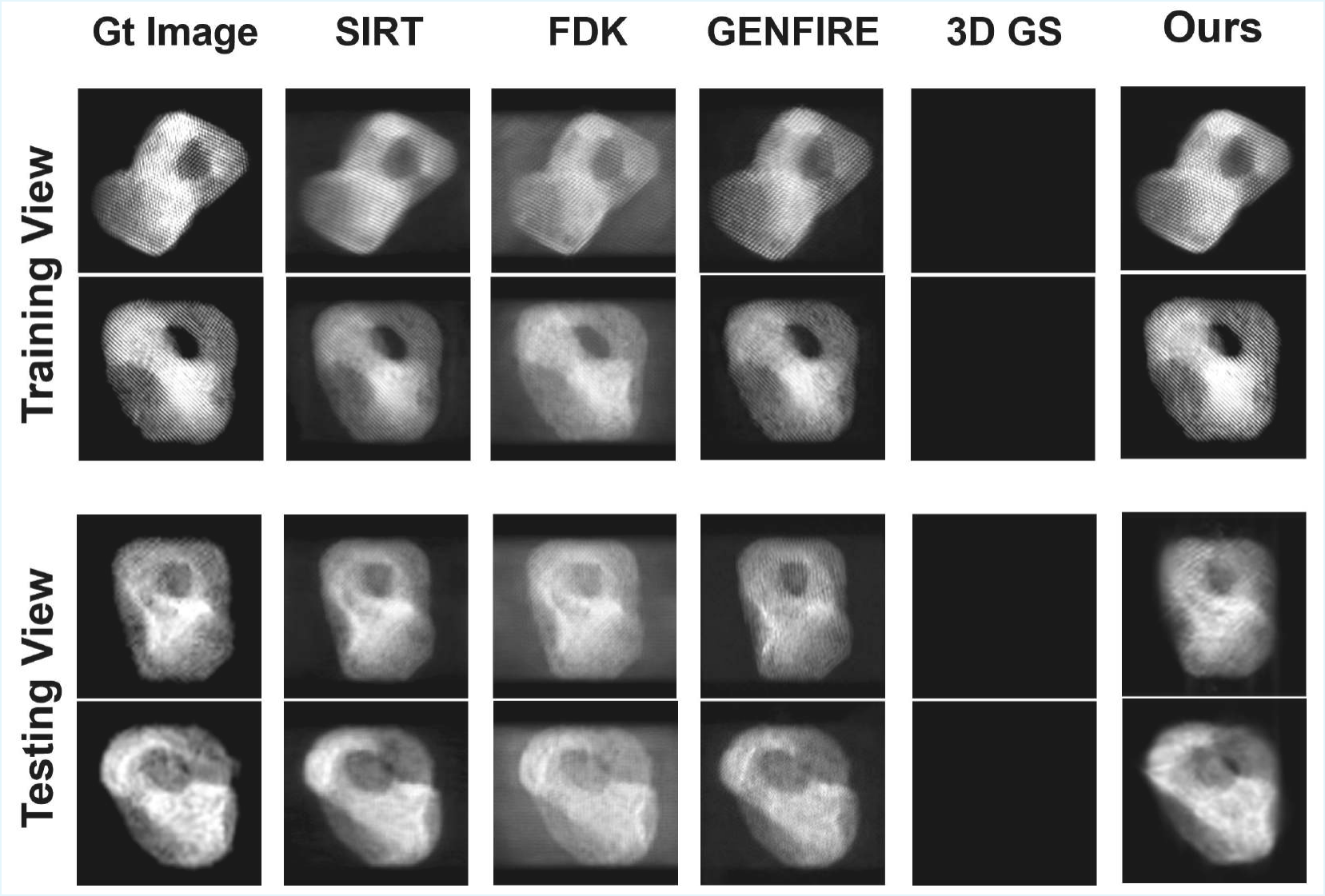}
  \caption{%
    Results of qualitative comparison under 45-view. We performed 3-D reconstruction of PtNi nanomaterial blocks from 45 tilt-series projections. Compared methods are SIRT~\cite{andersen1984simultaneous}, FDK~\cite{feldkamp1984practical}, GENFIRE~\cite{pryor2017genfire}, and 3D GS~\cite{kerbl20233d}.%
    }
  \label{fig:Comparison_45}
\end{figure}

\begin{table}[t]
\centering
\caption{Quantitative comparison under 45-view reconstruction.}
\begin{tabular}{lS[table-format=2.2]S[table-format=1.3]S[table-format=2.2]S[table-format=1.3]}
\toprule
& \multicolumn{2}{c}{Train view} & \multicolumn{2}{c}{Test view} \\
\cmidrule(lr){2-3}\cmidrule(lr){4-5}
Method & {PSNR$\uparrow$} & {SSIM$\uparrow$} & {PSNR$\uparrow$} & {SSIM$\uparrow$} \\
\midrule
SIRT~\cite{andersen1984simultaneous}            & 26.41 & 0.645 & 23.58 & 0.519 \\
FDK~\cite{feldkamp1984practical}             & 17.17 & 0.478 & 15.53 & 0.336 \\
GENFIRE~\cite{pryor2017genfire}         & 28.19 & 0.697 & 26.05 & 0.587 \\
3D GS~\cite{kerbl20233d}           &  8.93 & 0.201 &  8.93 & 0.201 \\
Ours & \textbf{33.02} & \textbf{0.885} & \textbf{29.17} & \textbf{0.790} \\
\bottomrule
\end{tabular}
\label{tab:quant_45}
\end{table}

To enhance perceptual alignment and preserve high-contrast edges such as core-shell interfaces, we employ an SSIM-based loss that quantifies structural similarity within sliding windows across each projection, \emph{i.e.},
\begin{equation}
\mathcal{L}_{\text{ssim}} = \frac{1}{N} 
\sum_{\theta=1}^{N} 
\left[ 1 - \text{SSIM}(I_{\text{render}}^\theta, I_{\text{meas}}^\theta; W) \right].
\end{equation}

A 3D total variation (TV) regularization term promotes structural smoothness in the voxelized volume by penalizing abrupt changes across neighboring voxels. Formally, the anisotropic TV loss is defined as
enforces structural smoothness in the voxelized volume $V$ via
\begin{align}
\mathcal{R}_{\text{3DTV}}
&= \sum_{(x,y,z)\in V}\bigl(
     \|\nabla_x V\|_1 + \|\nabla_y V\|_1 + \|\nabla_z V\|_1
   \bigr), 
\end{align}
where $\nabla_x V$, $\nabla_y V$, $\nabla_z V$ denote finite differences along the three spatial axes. This formulation can be intuitively interpreted as a clamped neighbor sum. For each voxel, directional differences with its adjacent neighbors are computed, clamped via absolute value to ignore sign, and aggregated with directional weights.

In our implementation, we extend this idea to a 3D 8-neighbor configuration, as illustrated in the right of Figure.\ref{fig:pipeline}, where each voxel interacts with its immediate neighbors along both axial and diagonal directions. The resulting regularization encourages local consistency while preserving sharp transitions, especially in sparsely sampled or noisy volumes.

We summarize the total loss formulation as
\begin{align}
\mathcal{L}_{\text{total}}
&= \lambda_{\text{pixel}}  \mathcal{L}_{\text{pixel}}
 + \lambda_{\text{freq}}   \mathcal{L}_{\text{freq}}\\
&+ \lambda_{\text{ssim}}   \mathcal{L}_{\text{ssim}}
 + \lambda_{\text{3DTV}}   \mathcal{R}_{\text{3DTV}}.
\end{align}

\section{Experiments}
In this section, we evaluate DenZa-Gaussian on real nanomaterial tilt series from Atomic-scale identification of active sites of oxygen reduction nanocatalysts~\cite{yang2024atomic}. Evaluation metrics are defined to assess reconstruction quality. We compare DenZa-Gaussian with traditional iterative and recent neural methods under two settings: (1) 45-view reconstruction with sufficient projections, and (2) 15-view sparse reconstruction for low-dose scenarios. These settings test adaptability to varying projection densities. We also perform discussions to assess the contribution of key components and input configurations.

\subsection{Experimental Settings.}
\noindent\textbf{Environment.}All experiments were conducted on a single NVIDIA RTX 3090 GPU.

\noindent\textbf{Datasets.}
Our dataset consists of ADF-STEM tilt series of PtNi and Mo-doped PtNi oxygen reduction nanocatalysts, originally published in \textit{Atomic-scale identification of active sites of oxygen reduction nanocatalysts}. Following standard 3DGS protocols, the 2D projections are utilized as our ground truth.~\cite{kerbl20233d} 
Each projection was binned to a resolution of 256×256 pixels. We divided the multi-view images into training and test subsets using two protocols, 45 projections representing adequate angular sampling and 15 projections yielding sparse angular sampling, enabling assessment of reconstruction quality under both standard and limited view conditions.

\noindent\textbf{Evaluation metrics.}
We adopt Peak Signal-to-Noise Ratio (PSNR)~\cite{jahne2005digital} and Structural Similarity Index (SSIM)~\cite{wang2004image} as core metrics for evaluating reconstruction quality. PSNR measures pixel-level fidelity by directly comparing the reconstructed 3D volume with the ground truth, where higher values indicate lower reconstruction error and better overall quality. SSIM emphasizes structural consistency and detail preservation. To compute it, we extract 2D slices from the reconstructed volume along axial, coronal, and sagittal planes, and average the SSIM scores across all slices. Higher SSIM values reflect stronger structural agreement with the ground truth.

\subsection{45-view reconstruction}
In this section, we present comprehensive experiments on 45-view reconstruction of PtNi nanocatalysts to validate the effectiveness of our method. The sufficient number of projections provides a well-conditioned setting that enables fair 

\begin{figure}[t]
  \centering
  \includegraphics[width=\columnwidth]{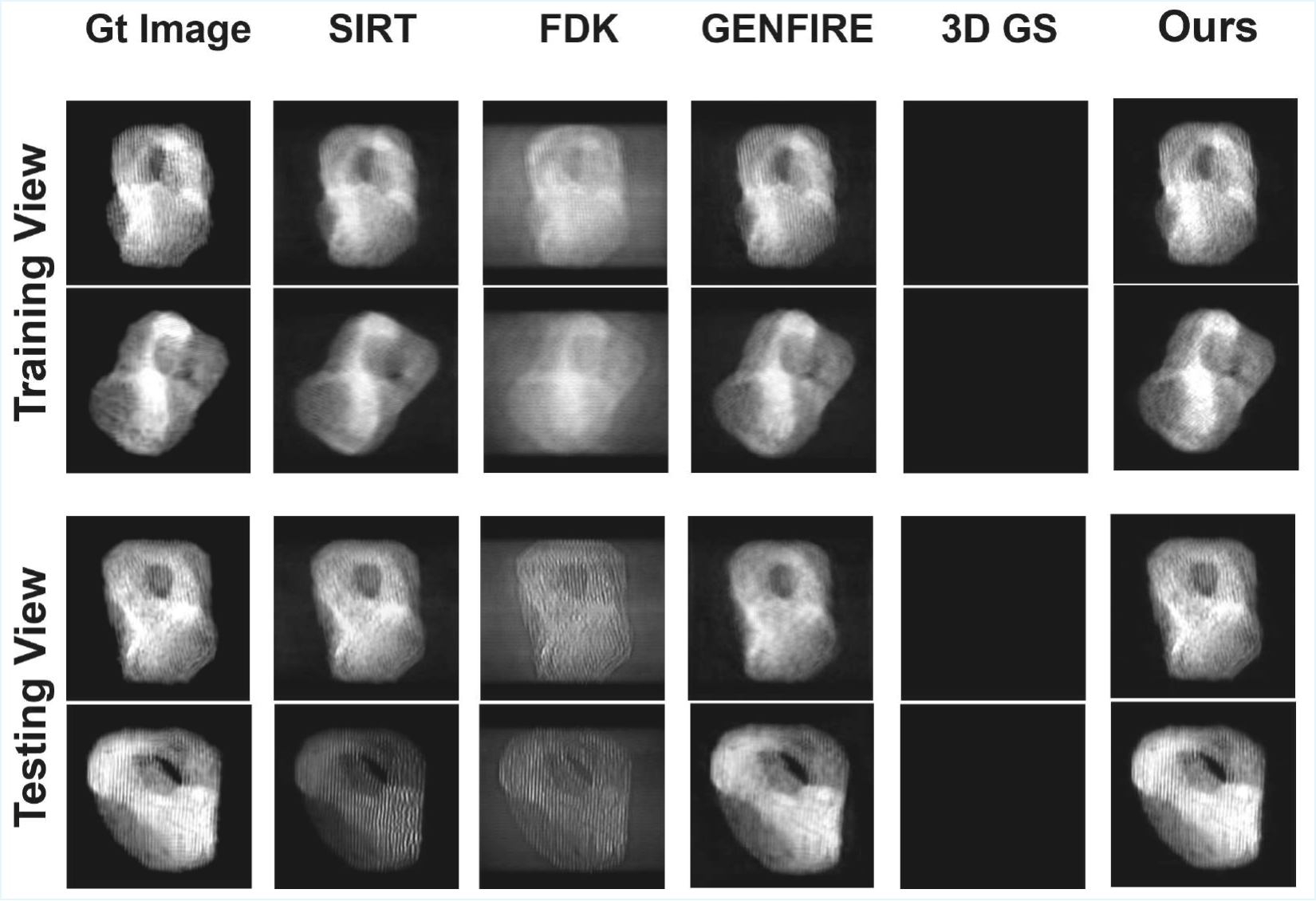}
  \caption{%
    Results of qualitative comparison under 15-view. We performed 3-D reconstruction of PtNi nanomaterial blocks from 15 tilt-series projections. Compared methods are SIRT~\cite{andersen1984simultaneous}, FDK~\cite{feldkamp1984practical}, GENFIRE~\cite{pryor2017genfire}, and 3D GS~\cite{kerbl20233d}.%
    }
  \label{fig:Comparison_15}
\end{figure}

\begin{table}[t]
\centering
\caption{Quantitative comparison under 15-view reconstruction.}
\begin{tabular}{lS[table-format=2.2]S[table-format=1.3]S[table-format=2.2]S[table-format=1.3]}
\toprule
& \multicolumn{2}{c}{Train view} & \multicolumn{2}{c}{Test view} \\
\cmidrule(lr){2-3}\cmidrule(lr){4-5}
Method & {PSNR$\uparrow$} & {SSIM$\uparrow$} & {PSNR$\uparrow$} & {SSIM$\uparrow$} \\
\midrule
SIRT~\cite{andersen1984simultaneous}  & 22.93 & 0.603 & 21.54 & 0.499 \\
FDK~\cite{feldkamp1984practical}      & 15.09 & 0.445 & 14.13 & 0.314 \\
GENFIRE~\cite{pryor2017genfire}       & 25.23 & 0.672 & 24.07 & 0.508 \\
3D GS~\cite{kerbl20233d}              &  8.93 & 0.201 &  8.93 & 0.201 \\
Ours & \textbf{31.73} & \textbf{0.798} & \textbf{27.53} & \textbf{0.751} \\
\bottomrule
\end{tabular}
\label{tab:quant_15}
\end{table}

\noindent comparison across methods. We benchmark our approach against traditional iterative algorithms and the neural rendering technique 3D Gaussian Splatting, assessing performance through both qualitative visualizations and quantitative metrics.

\noindent\textbf{Qualitative Comparison.}
As shown in Figure.\ref{fig:Comparison_45}, we present qualitative results of 45-view reconstruction, highlighting contour accuracy, artifact suppression, and intensity consistency with the ground truth. SIRT yields accurate global contours but introduces noticeable artifacts. FDK preserves overall outlines yet suffers from more pronounced distortions. GENFIRE produces sharper contours than our method but fails to capture the intensity relationships in projection images, resulting in brightness misalignment with the ground truth. 3D Gaussian Splatting fails to generate valid reconstructions, producing nearly black outputs. In contrast, our method preserves fine structural details and achieves consistent intensity alignment with the ground truth, demonstrating robust performance across both training and testing views.

\noindent\textbf{Quantitative Comparison.}
We quantify reconstruction performance using PSNR and SSIM. PSNR is reported because higher values indicate superior pixel-level fidelity. SSIM is reported because higher values indicate greater structural consistency, both metrics are evaluated on training and test views. As presented in Table~\ref{tab:quant_45}, traditional methods and 3D GS exhibit significant limitations across all metrics. SIRT achieves PSNR values of 26.41 for training view and 23.58 for testing view, with SSIM scores of 0.645 and 0.519 respectively. FDK performs worse, with PSNR as low as 17.17 for training view and 15.53 for testing view, and SSIM of 0.478 and 0.336. GENFIRE shows a relatively higher training view PSNR of 28.19 and 26.05 in the testing view, with SSIM values of 0.697 for training and 0.587 for testing revealing poor structural consistency. 3D GS fails to generate valid reconstructions, yielding the lowest PSNR of 8.93, and SSIM of 0.201 both for training and testing. Our method demonstrates substantial superiority, PSNR further improves to 33.02 for training and 29.17 for testing, and SSIM reaches 0.885 and 0.790. These results align closely with qualitative observations, confirming our method maintains high pixel-level fidelity and robust structural consistency, outperforming all compared methods in 45-view reconstruction.

\begin{table}[t]
\centering
\caption{Ablation on Scattering-View-Consistent coefficient $\gamma$.}
\begin{tabular}{lS[table-format=2.2]S[table-format=1.3]S[table-format=2.2]S[table-format=1.3]}
\toprule
& \multicolumn{2}{c}{Train view} & \multicolumn{2}{c}{Test view} \\
\cmidrule(lr){2-3}\cmidrule(lr){4-5}
Method & {PSNR$\uparrow$} & {SSIM$\uparrow$} & {PSNR$\uparrow$} & {SSIM$\uparrow$} \\
\midrule
w/o
& 10.32 & 0.321 & 9.15 & 0.253 \\
w & \textbf{31.73} & \textbf{0.798} & \textbf{27.53} & \textbf{0.751} \\
\bottomrule
\end{tabular}
\label{tab:ablation_scattering}
\end{table}

\subsection{15-view reconstruction}
In this section, we investigate the performance of each method under 15-view reconstruction settings for PtNi nanocatalysts. Given the limited number of projections, this scenario challenges the robustness of methods in handling data insufficiency.

\noindent\textbf{Qualitative Comparison.}
As shown in Figure.\ref{fig:Comparison_15}, we visualize the qualitative results of 15-view reconstruction, focusing on structural fidelity, artifact presence, and consistency between training and testing views. In the training view, SIRT appears noticeably blurry, failing to capture the fine details of the PtNi structure. In the testing view, it suffers from severe blurring, losing structural clarity. FDK exhibits substantial artifacts in both training and testing views, particularly in the testing view, the artifacts are prominent and accompanied by significant blurring, making the structure barely recognizable. 3D Gaussian Splatting fails to generate valid reconstructions, producing nearly black images, similar to its performance in the 45-view setting. In contrast, our method reconstructs the fine structural details, maintaining high consistency with the ground truth across both training and testing views, demonstrating robust performance even under the data-scarce 15-view conditions.

\noindent\textbf{Quantitative Comparison.}
As shown in Table~\ref{tab:quant_15}, SIRT achieves a training view PSNR of 22.93 and SSIM of 0.603, while its testing view performance drops to 21.54 in PSNR and 0.499 in SSIM, indicating limited robustness. FDK performs worse, with training view PSNR as low as 15.09 and SSIM of 0.445, and testing view PSNR further decreasing to 14.13 with SSIM of 0.314, reflecting poor pixel fidelity and structural consistency. 3D GS fails to generate valid reconstructions, yielding the lowest PSNR and SSIM across all metrics.
In contrast, our method shows significant superiority, reaching the highest training view PSNR of 31.73 and SSIM of 0.798, along with the best testing view PSNR of 27.53 and SSIM of 0.751. These quantitative results align with qualitative observations, confirming that our method maintains high pixel-level fidelity and structural consistency even under the data-scarce 15-view conditions, outperforming all compared methods.

\begin{table}[t]
\centering
\caption{Ablation on $Density \cdot Z^{\alpha}$.}
\begin{tabular}{lS[table-format=2.2]S[table-format=1.3]S[table-format=2.2]S[table-format=1.3]}
\toprule
& \multicolumn{2}{c}{Train view} & \multicolumn{2}{c}{Test view} \\
\cmidrule(lr){2-3}\cmidrule(lr){4-5}
Method & {PSNR$\uparrow$} & {SSIM$\uparrow$} & {PSNR$\uparrow$} & {SSIM$\uparrow$} \\
\midrule
Denza     & 31.02 & 0.885 & 28.17 & 0.790 \\
Density+Z & \textbf{31.20} & \textbf{0.890} & \textbf{28.30} & \textbf{0.797} \\
\bottomrule
\end{tabular}
\label{tab:ablation_param}
\end{table}

\subsection{3D Volume}
To further illustrate the superiority of our method in 3D volume reconstruction, we analyze the comparison figures in Figure.\ref{fig:3D_volume}\textcolor{cvprblue}{(a)}. The left group of images corresponds to the result of GENFIRE, which achieves the best performance in 3D volume reconstruction among comparison methods. While the right group belongs to our method. 

Visually, GENFIRE yields relatively uniform overall structures but suffers from notable noise, and atoms fail to be reconstructed as individual, distinct bright spots, leading to blurred details. In contrast, our method presents each atom as a single clear bright spot with significantly reduced noise, and the edges are much neater and smoother. This superior detail preservation and structural fidelity make our method more valuable in practical applications, such as atomic-scale structural analysis and characterization of nanomaterials, where precise and reliable structural information is crucial.

\subsection{Discussion}
\noindent\textbf{Scattering-View-Consistent coefficient $\boldsymbol{\Gamma}$.} We evaluated the proposed bias correction coefficient $\gamma$ through ablation experiments on 45-view ADF-STEM 3D reconstructions. Models with and without this correction were trained on PtNi and Mo-doped PtNi nanocatalyst tilt series. Reconstruction quality was assessed using PSNR and SSIM across training and test views. As shown in Table~\ref{tab:ablation_scattering}, the baseline without correction achieved only 10.32 PSNR and 0.321 SSIM on training views, and 9.15 PSNR and 0.253 SSIM on test views. The corrected model significantly improved these metrics to 31.73 PSNR and 0.798 SSIM for training, and 27.53 PSNR and 0.751 SSIM for testing. These results demonstrate that our proposed correction effectively mitigates the severe reconstruction quality degradation caused by scattering integration bias.

\begin{figure}[t]
  \centering
  \includegraphics[width=\linewidth]{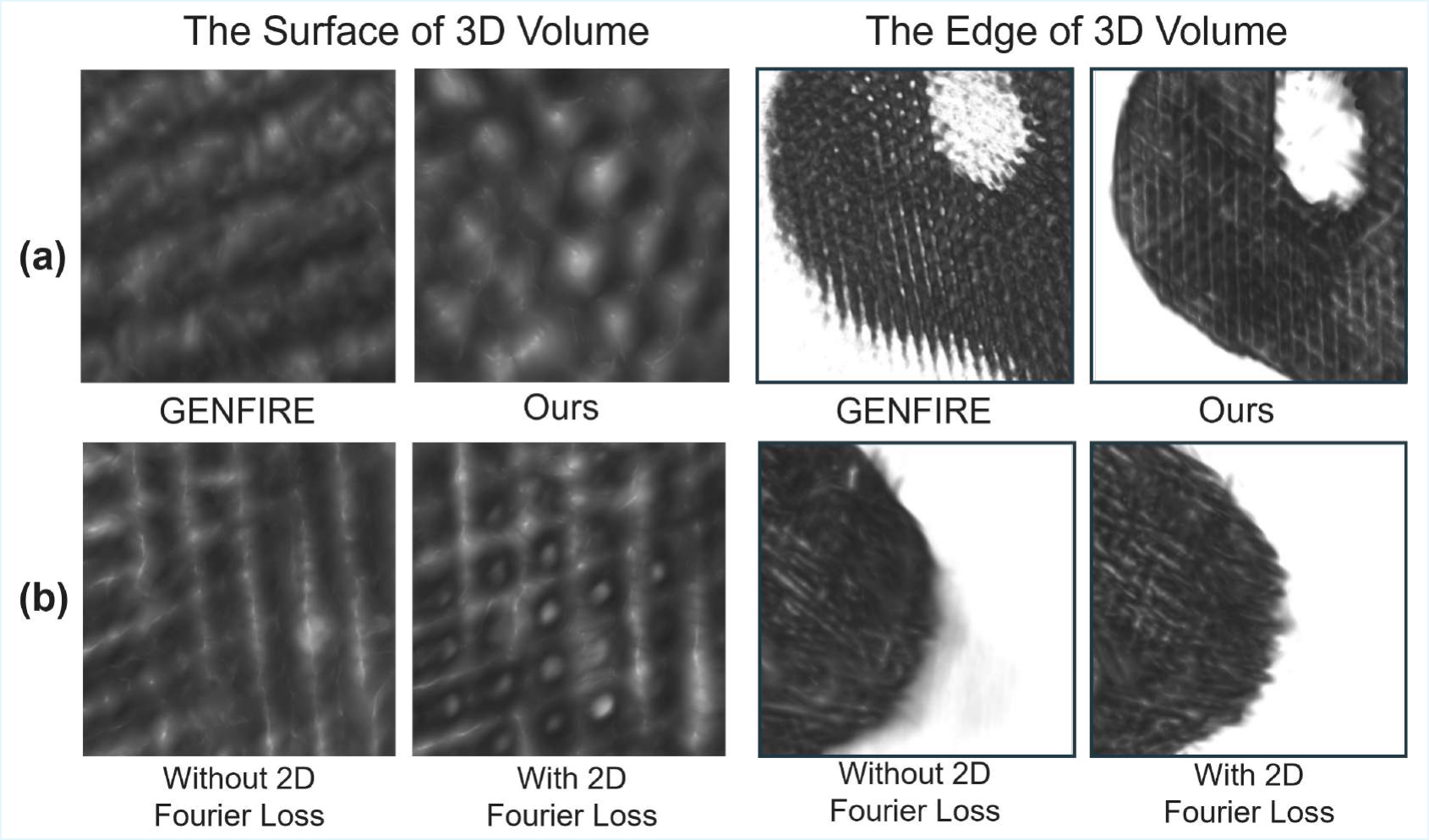}
  \caption{Comparison of 3D volume reconstruction results. The left panel compares the reconstructed 3D volume surfaces, while the right panel focuses on edge reconstruction. Part (a) presents a comparison between GENFIRE and our method on full 3D volumes, and part (b) shows an ablation study of the 2D Fourier loss under sparse-view conditions.}
  \label{fig:3D_volume}
\end{figure}

\noindent\textbf{2D Fourier Loss.} Figure.\ref{fig:3D_volume}\textcolor{cvprblue}{(b)} presents ablation experiments on the 2D Fourier loss under 15-view projection. The figure compares reconstructed 3D volume surfaces and edges without and with this loss. A clear contrast emerges: surfaces reconstructed with the Fourier loss retain distinct atomic bright spots absent in the baseline. Meanwhile, edges utilizing the Fourier loss appear sharp and smooth, whereas those without exhibit severe artifacts. These observations confirm that our 2D Fourier loss effectively preserves high-frequency details and suppresses missing-wedge artifacts under sparse-view conditions.

\noindent\textbf{$\boldsymbol{Density \cdot Z^{\alpha}}$.} We validate simplifying ${Density \cdot Z^{\alpha}}$ into a single denza parameter by comparing it against a variant learning density and $Z$ separately. While Table~\ref{tab:ablation_param} shows this separated configuration yields slightly higher metrics, its learned $Z$ values deviate from true Pt and Ni atomic numbers. This indicates the marginal gains stem from mathematical flexibility rather than physical accuracy. We discard this setup to preserve physical validity and consistency with ADF-STEM principles. Future work will explore multimodal imaging constraints to reliably estimate $Z$.

\begin{figure}[t]
  \centering
  \includegraphics[width=\linewidth]{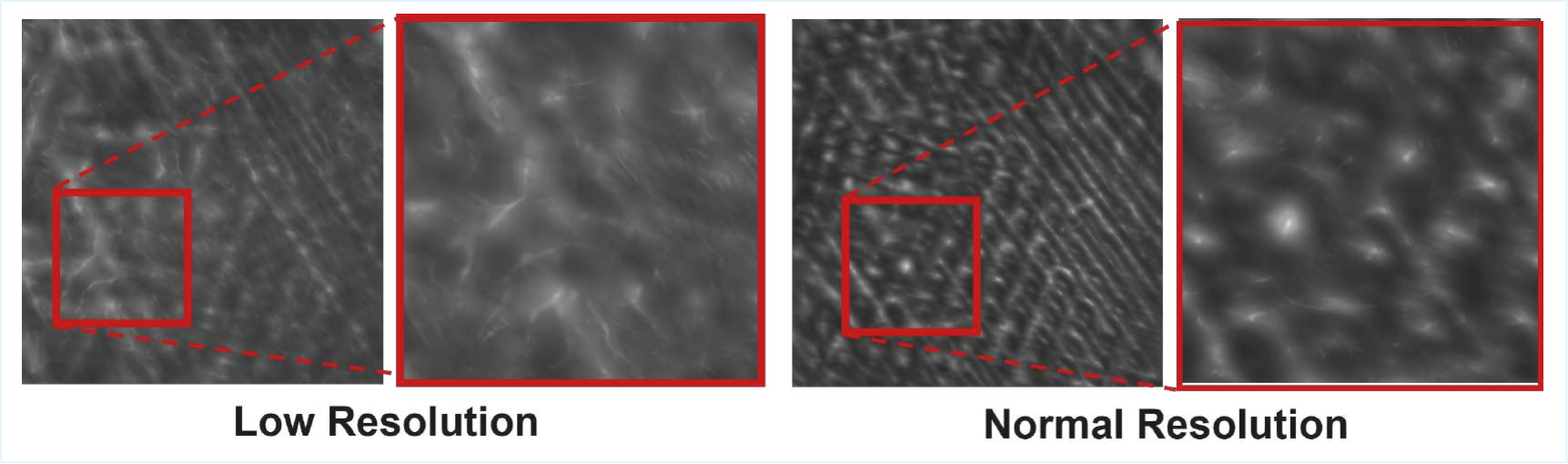}
  \caption{Impact of ground truth (GT) image resolution. In each block, the left panel shows the surface of the reconstructed 3D volume using input images downsampled by merging every four pixels into one, while the right panel corresponds to reconstruction from the original resolution input.}
  \label{fig:Discussion_1}
\end{figure}

\noindent\textbf{The impact of GT images.} Figure~\ref{fig:Discussion_1} compares 3D reconstructions using normal-resolution inputs against downsampled versions merging four pixels into one. The downsampled results appear noticeably smoother and lack fine details due to the irreversible loss of high-frequency information during pixel merging. This demonstrates that high-frequency content in ground truth images is essential for preserving structural precision in high-fidelity 3D reconstructions. Future work will address this degradation by enhancing high-frequency features via input preprocessing and incorporating explicit frequency-domain constraints, like our 2D Fourier amplitude loss, to further improve detail fidelity.

\section{Conclusion}
We propose DenZa-Gaussian, an approach tailored for sparse-view ADF-STEM tomography reconstruction that addresses the tradeoff between sample preservation and reconstruction fidelity. By adapting 3D Gaussian Splatting to ADF-STEM physical imaging mechanisms, we introduce the learnable denza coefficient, Scattering-View-Consistent efficient $\gamma$, and a composite loss combining pixel-wise, frequency-domain, and 3D TV regularization. This design resolves physical mismatches and suppresses missing wedge artifacts. Experiments on 45-view and 15-view tilt series show that DenZa-Gaussian consistently outperforms traditional and neural baselines in both quantitative metrics and qualitative fidelity, preserving fine nanoscale details even under sparse-view conditions. This work advances ADF-STEM tomography for dose-sensitive nanomaterials and lays the foundation for future improvements in high-frequency enhancement and multi-modal integration.

\clearpage
\section*{Acknowledgements}
This work is supported by the National Natural Science Foundation of China (62331006), and the Fundamental Research Funds for the Central Universities.
{
    \small
    \bibliographystyle{ieeenat_fullname}
    \bibliography{main}
}


\end{document}